\documentclass[11pt,a4paper]{article}
\usepackage[hyperref]{eacl2021}
\usepackage{times}
\usepackage{latexsym}

\usepackage{url}
\usepackage{multirow}
\usepackage{booktabs}
\usepackage{tabularx}
\usepackage{graphics}
\usepackage{graphicx}
\usepackage{amssymb} 

\definecolor{berry}{HTML}{BC3754}

\usepackage{microtype}

\aclfinalcopy 
\def\aclpaperid{192} 

\title{FEWS: Large-Scale, Low-Shot Word Sense\\ Disambiguation with the Dictionary}

\author{Terra Blevins, Mandar Joshi, and Luke Zettlemoyer \\
        Paul G. Allen School of Computer Science \& Engineering, University of Washington \\
        Facebook AI Research, Seattle \\
        {\tt \{blvns, mandar90, lsz\}@cs.washington.edu}}

\date{}

\begin{document}
\maketitle
\begin{abstract}
Current models for Word Sense Disambiguation (WSD) struggle to disambiguate rare senses, despite reaching human performance on global WSD metrics. This stems from a lack of data for both modeling and evaluating rare senses in existing WSD datasets. In this paper, we introduce FEWS (Few-shot Examples of Word Senses), a new low-shot WSD dataset automatically extracted from example sentences in Wiktionary. FEWS has high sense coverage across different natural language domains and provides: (1) a large training set that covers many more senses than previous datasets and (2) a comprehensive evaluation set containing few- and zero-shot examples of a wide variety of senses. We establish baselines on FEWS with knowledge-based and neural WSD approaches and present transfer learning experiments demonstrating that models additionally trained with FEWS better capture rare senses in existing WSD datasets. Finally, we find humans outperform the best baseline models on FEWS, indicating that FEWS will support significant future work on low-shot WSD.
\end{abstract}

\begin{table*}[t]
    \centering
    \small
    \begin{tabularx}{0.85\textwidth}{l | r r r r r r} 
    \toprule
    \textbf{Data Split} & \textbf{\# Examples}& \textbf{\# Tokens} & \textbf{\# Annot.} & \textbf{\# Sense Types} & \textbf{\# Word Types} & \textbf{Ambiguity} \\
    \hline 
    Overall & 121,459 & 3,259,240 & 131,278 & 71,391 & 35,416 & 5.62 \\
    \hline
    Train & 87,329 & 2,551,358 & 96,023 & 52,928 & 30,450 & 4.98 \\
    Ext. Train & 101,459 & 2,683,345 & 111,278 & 61,391 & 31,937 & 5.71 \\
    \hline
    Dev & 10,000 & 287,673 & 10,000 & 10,000 & 8,682 & 5.09 \\
    \hspace{3mm}Few-shot & 5,000 & 149,791 & 5,000 & 5,000 & 4,417 & 4.77 \\
    \hspace{3mm}Zero-shot & 5,000 & 137,882 & 5,000 & 5,000 & 4,661 & 5.41 \\
    \hline
    Test & 10,000 & 288,222 & 10,000 & 10,000 & 8,709 & 5.10 \\
    \hspace{3mm}Few-shot & 5,000 & 149,384 & 5,000 & 5,000 & 4,449 & 4.71 \\
    \hspace{3mm}Zero-shot & 5,000 & 138,838 & 5,000 & 5,000 & 4,666 & 5.49 \\
    \toprule
    \end{tabularx}
    \caption{FEWS data statistics. The development and test sets are balanced across senses and split evenly between few-shot examples (with support in the training set) and zero-shot examples. The extended training set (Ext. Train) adds short examples written by Wiktionary editors as additional training data.}
    \label{data-stats-table}
\end{table*}

\section{Introduction}
Word Sense Disambiguation (WSD) is the task of identifying the \textit{sense}, or meaning, that an ambiguous word takes in a specific context. Recent WSD models~\cite{huang2019glossbert, blevins2020moving, bevilacqua2020breaking} have made large gains on the task, surpassing the estimated 80\% F1 human performance upper bound on WordNet annotated corpora \cite{navigli2009word}.
Despite this breakthrough, the task remains far from solved: performance on rare and zero-shot senses is still low, and in general, current WSD models struggle to learn senses with few training examples \cite{kumar2019zero, blevins2020moving}. This performance gap stems from limited data for rare senses in current WSD datasets, which are annotated on natural language documents that contain a Zipfian distribution of senses \cite{postma2016more}. 

\begin{figure}
\centering
\begin{tabular}{c l}
 \toprule
\multicolumn{2}{l}{\textbf{C}: I \textbf{liked} my friend's last status...} \\
 & \textbf{S1}: to enjoy... [or] be in favor of.\\
 \checkmark & \textbf{S2}: To show support for, or approval of, \\ & something on the Internet by marking it \\ & with a vote.\\
    \hline
 \multicolumn{2}{l}{\textbf{C}: A transistor-diode \textbf{matrix} is composed of} \\ \multicolumn{2}{l}{vertical and horizontal wires with a transistor} \\  \multicolumn{2}{l}{at each intersection.} \\
 \checkmark &\textbf{S1}: A grid-like arrangement of electronic\\ & components, especially one intended for \\ & information coding, decoding or storage. \\
 &\textbf{S2}: A rectangular arrangement of \\ & numbers or terms having various uses \\  & $[$in mathematics$]$.  \\
 \toprule
\end{tabular}
\caption{Sample contexts (\textbf{C}) from FEWS with ambiguous words and a subset of candidate sense definitions (\textbf{S}). FEWS covers a wide range of senses, including new senses and domain-specific senses.}
\label{teaser-fig}
\end{figure}

More generally, since each word has a different set of candidate senses and new senses are regularly coined, it is almost impossible to gather a large number of examples for each sense in a language. This makes the few-shot learning setting particularly important for WSD. We introduce \textbf{FEWS} (Few-shot Examples of Word Senses), a dataset built to comprehensively train and evaluate WSD models in few- and zero-shot settings. Overall, the contributions of FEWS are two-fold: as training data, it exposes models a broad array of senses in a low-shot setting, and the large evaluation set allows for more robust evaluation of rare senses.

FEWS achieves high coverage of rare senses by automatically extracting example sentences from Wiktionary definitions. Wiktionary is an apt data source for this purpose, containing examples for over 71,000 senses (Table \ref{data-stats-table}). Not only is this sense coverage higher than existing datasets (e.g., SemCor, the largest manually annotated WSD dataset, only covers approximately 33,000 senses~\cite{miller1993semantic}), it also extends to senses related to new domains (Figure \ref{teaser-fig}). 

We establish performance baselines on FEWS with both knowledge-based approaches and a recent neural biencoder model for WSD~\cite{blevins2020moving}. We find that the biencoder, despite being the strongest baseline on FEWS, still underperforms human annotators, particularly on zero-shot senses where the biencoder trails by more than 10\%. We also present transfer learning experiments and find adding FEWS as additional training data improves performance on all but the most frequent senses (MFS) in the WSD Evaluation Framework~\cite{raganato2017word}; this suggests that future improvements on FEWS could generalize other WSD benchmarks. FEWS is available at \url{https://nlp.cs.washington.edu/fews}.

\section{Related Work}

WSD is a long-standing NLP task and is the focus of many datasets. The current \textit{de facto} benchmark for modeling English WSD is the WSD Evaluation Framework \cite{raganato2017word}, which includes the SemCor dataset \cite{miller1993semantic} as training data and consolidates a number of evaluation sets \cite{pradhan2007semeval, palmer2001english, snyder2004english, navigli2013semeval, moro2015semeval} into a standardized evaluation suite. These datasets are annotated with the senses (known as \textit{synsets}) from Wordnet, a manually constructed ontology of semantic relations \cite{miller1993semantic}. 

Most existing datasets for WSD, including those in the WSD Evaluation framework and others like \citet{pradhan2007ontonotes}, are annotated on natural language documents that contain a Zipfian distribution of word senses \cite{kilgarriff2004dominant}. This data source causes these datasets to have low coverage of rare senses, leading to worse performance on these less common senses \cite{postma2016more, kumar2019zero}. In contrast, we use examples sentences from Wiktionary as an alternative source of text for WSD data with FEWS. This means that FEWS has a more uniform sense distribution, providing more balanced coverage across different senses of words.

Wiktionary has previously been used as a resource for WSD research. Most work has investigated mapping Wiktionary senses onto WordNet synsets \cite{meyer2011psycholinguists, matuschek2013tacl}; other work has learned similar mappings for Wikipedia articles (\citet{mihalcea2007using, navigli2012babelnet}; inter alia). More similar to our work, \citet{henrich2012webcage} and \citet{segonne2019using} mine WSD examples from Wiktionary to augment labeled WSD data for non-English languages. However, FEWS is the first dataset specifically designed to evaluate zero and few-shot learning with the balanced dictionary sense distribution.

\begin{figure*}[t]
\centering
\includegraphics[width=\linewidth]{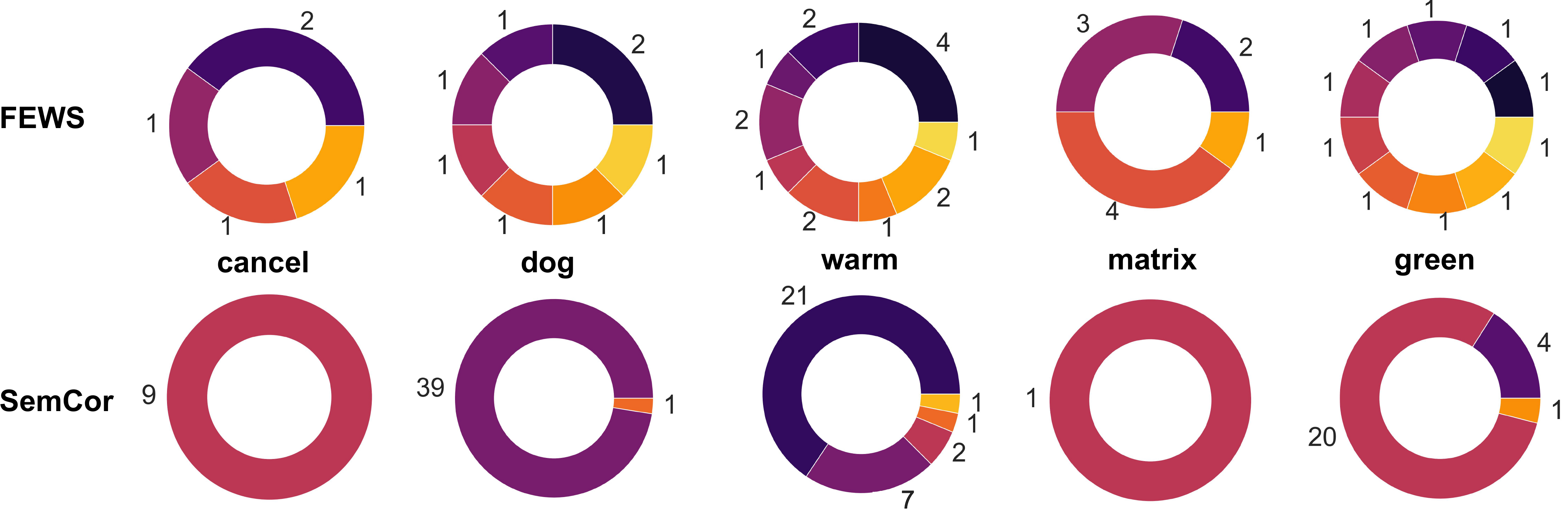}
\caption{Comparison of sense coverage for five words in the Semcor and FEWS training corpuses.}
\label{sense-coverage-fig}
\end{figure*}

\section{FEWS: Low-shot Learning for WSD}

FEWS (Few-shot Examples of Word Senses) is a new dataset for learning to do low-shot WSD.\footnote{We use the term \textit{low-shot} as an umbrella term for few- and zero-shot learning.} It is created with example contexts drawn from Wiktionary, an online collaborative dictionary.\footnote{https://en.wiktionary.org/} Since Wiktionary is curated by volunteers, the data is manually annotated and high quality, and there is no additional annotation cost to construct FEWS. Furthermore, using example contexts from a dictionary allows FEWS to cover many senses underrepresented in existing WSD datasets, such as rare senses of words or senses pertaining to specific domains. However, we note that due to the crowd-sourced nature of the examples in Wiktionary and the subjectivity of fine-grained sense distinctions, inconsistencies in the underlying data may introduce some annotation errors into FEWS.

Examples of the data in FEWS are shown in Figure \ref{teaser-fig}. In FEWS, each example context contains one or more instances of the ambiguous \textit{target word} and is labeled with the sense (and corresponding definition) that describes that word as used in the context; this is in contrast to \textit{all-words WSD}, where many of the content words in the context are annotated.

\subsection{Dataset Creation}
To create FEWS, we extracted all of the definitions for content words (nouns, verbs, adjectives, and adverbs) and example contexts associated with those definitions from a checkpointed version of English Wiktionary.\footnote{We focus English senses and filter out definitions for words in other languages; however, this data collection process could be expanded to other languages in Wiktionary.}
While processing the Wiktionary data, we collected two types of contexts: (1) \textit{quotations} (93\% of extracted examples), which are quotations of natural language text found by Wiktionary contributors that contain the target word used with the relevant sense, and (2) \textit{illustrations} (7\%), which are short sentences or fragments written by contributors to illustrate the word sense in context. The target words in each context are marked by the Wiktionary formatting metadata; examples where no words are marked or the marked word differs too much from the base form of the dictionary entry are discarded.\footnote{We define a marked word as too different from the base form if the longest common subsequence between them is $< 50\%$ of the length of the marked word.} We additionally filter out examples that are too short to provide a meaningful context for the marked word.

We then labeled the target words in each extracted context with the sense ID generated for the definition associated with that sentence; this gave us 254,506 annotated WSD example contexts covering 148,333 senses. Finally, we filtered out examples with monosemous words since predicting the sense in these cases is a trivial task. However, \citet{loureiro2020neglect} recently found that unambiguous examples can improve WSD performance; therefore, we include the unambiguous cases as an additional file in the FEWS dataset. After filtering these unambiguous examples from the main dataset, FEWS in total contains 121,459 examples covering 71,391 sense types. 

Finally, we split the data into training and evaluation sets. The majority of the data are \textit{quotations}, which we use to populate the train, development, and test sets as they more closely resemble naturally occurring text than the \textit{illustrations}. To create the development and test sets, we randomly select 10,000 examples for each evaluation set and ensure that each of these evaluation examples pertains to a different sense. We verify that half of those examples were labeled with senses that only occurred once in the unsplit data to create a zero-shot subset of each evaluation set, and the other half of the evaluation senses comprise the few-shot evaluation subsets. The remaining quotations that were not used for the development or test set are included as the training data. Finally, we remove the \textit{illustrations} for senses in the zero-shot evaluation subsets and add the remaining \textit{illustrations} to the training data; this addition makes the extended train set.

\begin{table}[t]
\centering
\begin{tabular}{l|r|r}
\toprule
\textbf{Eval Split} & \textbf{FEWS} & \textbf{WSD Fr.}\\
\hline
Dev & 10,000 & 375  \\
\hspace{3mm}Few-shot$\dagger$ & 4,529 & 67 \\
\hspace{3mm}Zero-shot & 5,000 & 50 \\
\hline 
Test & 10,000 & 3,669 \\
\hspace{3mm}Few-shot$\dagger$ & 4,603 & 761 \\
\hspace{3mm}Zero-shot & 5,000 & 796 \\
\toprule
\end{tabular}
\caption{The number of senses covered in the \textbf{FEWS} and \textbf{WSD Framework} evaluation sets. $\dagger$ To fairly compare against the WSD Framework, we only count few-shot examples as those have three or fewer supporting examples in their respective train set.}
\label{eval-stats-table}
\end{table}

\subsection{Dataset Analysis}
\label{dataset-analysis}

We present a comprehensive analysis of FEWS to demonstrate that the dataset provides high coverage of many diverse words senses in a low-shot manner.

\paragraph{High Coverage of Words and Senses} 
The FEWS dataset covers 35,416 polysemous words and 71,391 senses (Table \ref{data-stats-table}). The complete dataset covers 53.21\% of the senses for words that appear in it (out of their Wiktionary sense inventories). 
Figure \ref{sense-coverage-fig} shows this high coverage of senses for five different words compared to the coverage of the same words in SemCor \cite{miller1993semantic}. We see that while SemCor tends to have more examples for these common words, most examples correspond to a single sense of the word. However, the FEWS train set covers many more senses per word, albeit with fewer total examples.

This high coverage of senses is particularly important for the evaluation sets provided in FEWS (Table \ref{eval-stats-table}). Each evaluation set (development and test) covers 10,000 different senses; half of these examples are few-shot and occur in the training set, and the other half of the evaluation senses are zero-shot. In comparison, the current benchmark for WSD evaluation \cite{raganato2017word} only contains 796 unique zero-shot and 761 unique few-shot senses (where the sense is seen three or fewer times in the SemCor \cite{miller1993semantic} training set) across development and test evaluation sets. This much larger sample of few- and zero-shot evaluation examples means that FEWS provides a robust setting to evaluate model performance on less common senses.

\begin{figure}[t]
\centering
\includegraphics[width=0.8\linewidth]{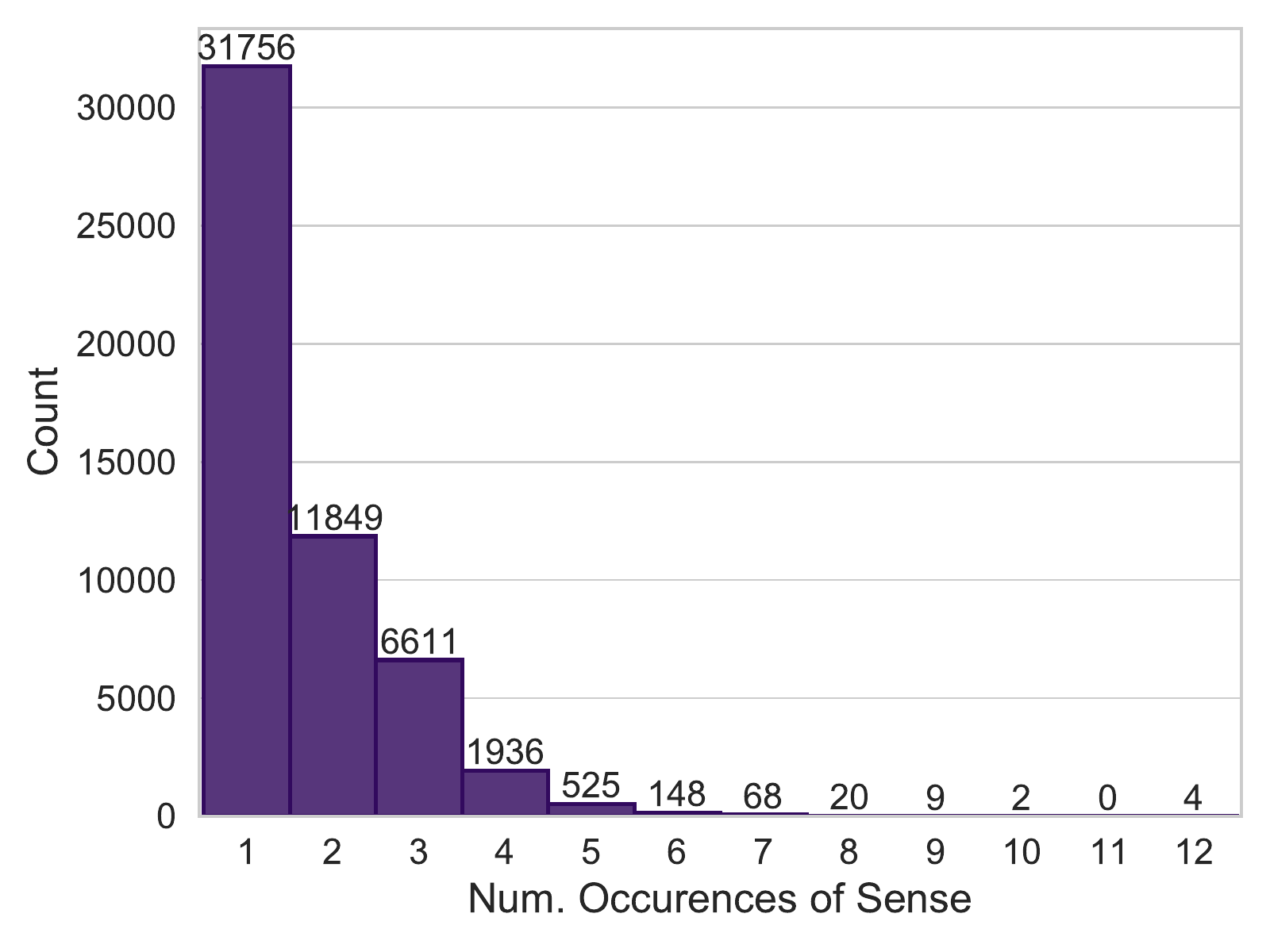}
\caption{Histogram of sense frequencies in the FEWS training data.}
\label{fewshot-hist}
\end{figure}

\paragraph{Low-Shot Learning}
Because the data in FEWS come from example sentences for definitions in Wiktionary, each sense occurs in only a few labeled examples. This low-shot nature of the data is shown for five common words in Figure \ref{sense-coverage-fig}: each sense of these words occurs only one to four times in the training data. Figure \ref{fewshot-hist} shows a histogram of the number of examples per sense in the full training set; we see that the majority of senses seen during training (60\%) only occur once and that there are on average 1.65 examples per sense. 

This set up also means that all evaluations on FEWS are low-shot (or zero-shot): senses in the few-shot development split have on average 2.06 supporting examples in the training set, with a maximum of 12 examples (these counts increase to 2.13 and 13, respectively, with the extended training data).

\begin{figure}[b!]
\centering
\includegraphics[width=0.7\linewidth]{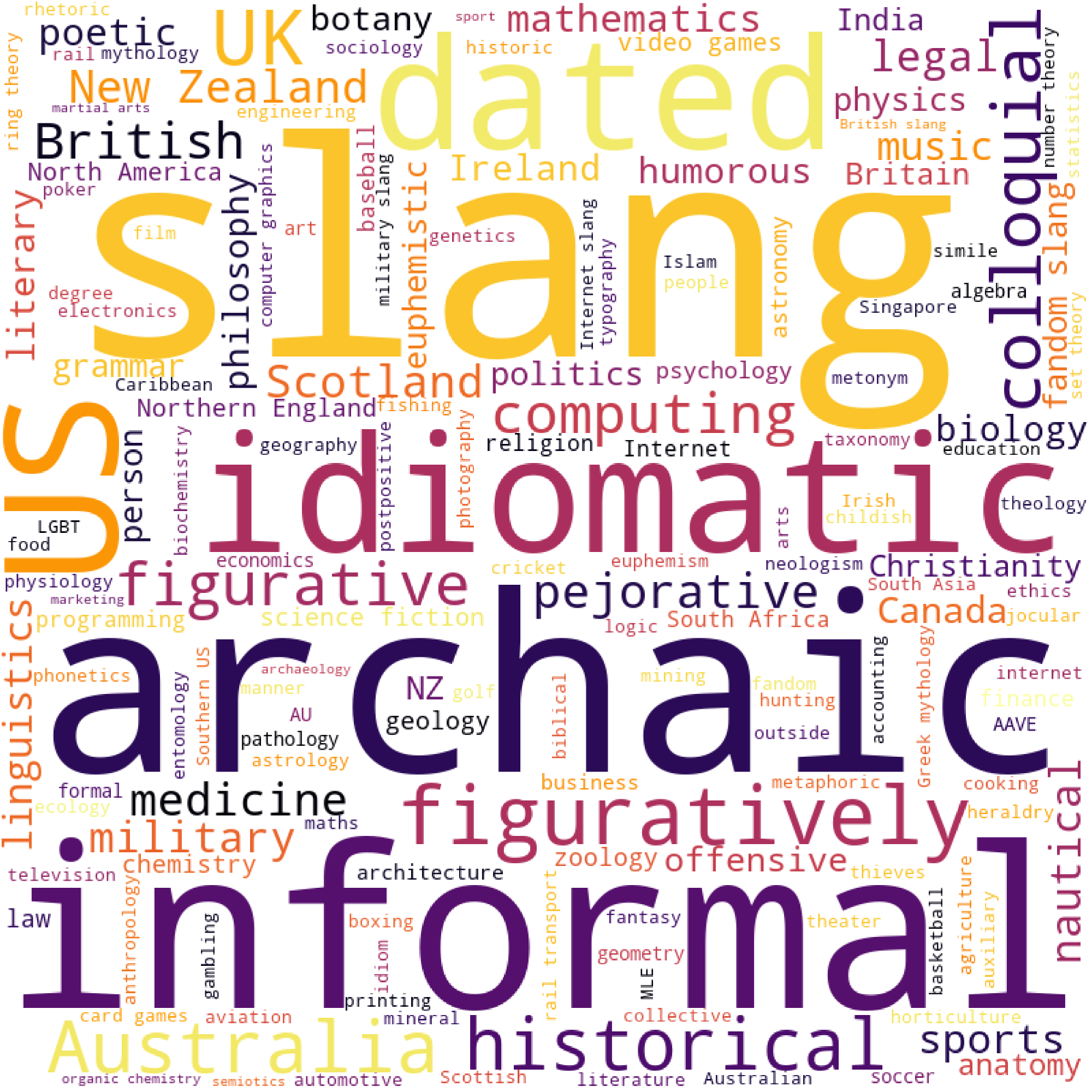}
\caption{Word cloud of the 300 most common tags in the FEWS senses. For clarity, we manually filter out syntactic and uninformative tags.}
\label{domain-wc}
\end{figure}

\paragraph{Domains in FEWS} Definitions in Wiktionary are tagged with keywords, which we include as metadata for their respective senses in the dataset. These keywords indicate that the senses in FEWS pertain to topics ranging from literature and archaic English to sports and the sciences and come from various English dialects (Figure \ref{domain-wc}). FEWS also covers many new domains not covered in existing WSD corpora, with six keywords corresponding to internet culture.

We also find that a number of the senses ($<$1\% in FEWS) are indicated to be toxic or offensive language: our analysis contains tags such as \textit{ethnic slurs}, \textit{offensive}, \textit{vulgar}, and \textit{derogatory} that correspond to examples of toxic language. For many of these examples, the meaning of the target word is only toxic due to the context in which it appears. These examples provide an opportunity for improving models for hate speech detection and related tasks, but we leave this exploration to future work.

\begin{table*}[t]
    \centering
    \begin{tabular}{l | r r r | r r r} 
    \toprule
     & \multicolumn{3}{c|}{\textbf{Dev}} & \multicolumn{3}{c}{\textbf{Test}} \\
     & \textbf{Full Set}& \textbf{Few-shot} & \textbf{Zero-shot} & \textbf{Full Set} & \textbf{Few-shot} & \textbf{Zero-shot} \\
    \hline
    \hline
    \multicolumn{7}{l}{\textbf{Knowledge-based baselines}} \\
    \hline
    MFS & 26.39 & 52.78 & 0.00 & 25.73 & 51.46 & 0.00 \\
    Lesk & 39.24 & 42.54 & 35.94 & 39.07 & 40.94 & 37.20 \\
    Lesk+emb & 42.54 & 44.94 & 40.14 & 41.53 & 44.08 & 38.98 \\
    \hline
    Human$\dagger$ & 80.11 & 80.44 & 79.87 & -- & -- & -- \\
    \hline
    \hline
    \multicolumn{7}{l}{\textbf{Neural baselines}} \\ 
    \hline
    Probe$_{BERT}$ & 36.17 & 72.34 & 0.00 & 36.07 & 72.14 & 0.00 \\
    BEM$_{BERT}$ & 73.81 & 79.28 & 68.34 & 72.77 & 79.06 & 66.48 \\
     \hline
    BEM$_{SemCor}$ & 74.36 & 79.72 & 69.00 & 72.98 & 78.88 & 67.08 \\
    BEM$_{zero-shot}\ddagger$ & 58.05 &	58.38 & 57.72 &	57.39 & 57.94 & 56.84 \\
    \toprule
    \end{tabular}
    \caption{Accuracy (\%) of our baselines (Section \ref{baselines-section}) and transfer learning models (Section \ref{transfer-section}) on the FEWS evaluation sets. Human performance ($\dagger$) is calculated on a subset of the development set and acts as an estimated upper bound on performance. \textbf{Probe$_{BERT}$} and \textbf{BEM$_{BERT}$} are baselines trained on FEWS; \textbf{BEM$_{SemCor}$} is a transfer learning model finetuned on SemCor before training on FEWS.  \textbf{BEM$_{zero-shot}$} ($\ddagger$) is a zero-shot transfer experiment in which the BEM trained on SemCor is evaluated on FEWS \textit{without} finetuning on the FEWS train set.}
    \label{fews-results-table}
\end{table*}

\section{Baselines for FEWS}
\label{baselines-section}
We run a series of baseline approaches on FEWS to demonstrate how current methods for WSD perform on this dataset. We consider a number of knowledge-based approaches (\textbf{Lesk}, \textbf{Lesk+emb}, and \textbf{MFS}) and two neural models that build on pretrained encoders (\textbf{Probe$_{BERT}$} and \textbf{BEM$_{BERT}$}). We also ascertain how well humans perform on FEWS as a potential upper bound for model performance.

\subsection{Knowledge-based Baselines}

\paragraph{Most Frequent Sense (MFS)} The MFS baseline assigns each target word in the evaluation with their candidate sense that is most frequently observed as the correct sense of that word in the training set. The MFS heuristic is known to be a particularly strong baseline in WSD datasets labeled on natural language documents \cite{kilgarriff2004dominant}; however, we expect this to be a weaker baseline on FEWS since the distribution of senses is much more balanced (and half of the evaluation senses are completely unseen during training).

\paragraph{Lesk} The simplified \textbf{Lesk} algorithm assigns to each ambiguous target word the sense whose gloss has the highest word overlap with the context surrounding that target word \cite{kilgarriff2000framework}. We specifically use the Lesk-definitions baseline from this work, meaning that we do \textit{not} include words from example sentences in the set compared against the context -- since these example sentences are used as the contexts in FEWS.

\paragraph{Lesk+emb} This baseline is an extension of the above approach that incorporates word embeddings \cite{basile2014enhanced}. A vector representation is built for the context around an ambiguous word ($v_c$) and the glosses of each sense of that word ($v_g$), where $v_c$ and $v_g$ are the element-wise sums of the word vectors for words in the context and gloss, respectively. The sense that corresponds to the $v_g$ with the highest cosine similarity to $v_c$ is chosen as the label for the target word. We use Glove embeddings \cite{pennington2014glove} for our implementation of this baseline. 
 
\subsection{Neural Baselines} 
\paragraph{Probe$_{BERT}$} This baseline is a linear classifier trained on contextualized representations output by the final layer of a frozen pretrained model; we use BERT as our pretrained encoder \cite{devlin2019bert}. We train this classifier by performing a softmax over all of the senses in the Wiktionary sense inventory and mask out any senses not relevant to the target word.

\paragraph{\textbf{BEM}} Our other neural baseline is the biencoder model (BEM) for WSD introduced by \citet{blevins2020moving}. The BEM has two independent encoders, a \textit{context encoder} that processes the context sentence (including the target word) and a \textit{gloss encoder} that encodes the glosses of senses into a sense representation. The BEM takes the dot product of the target word representation from the \textit{context encoder} and sense representations from the gloss encoder, and it labels the target word with the sense that has the highest dot product score. We train \textbf{BEM$_{BERT}$} by initializing each encoder with BERT and training on the FEWS train set.

\subsection{Human Performance} Finally, we calculate the estimated human performance on the FEWS development set. The three human annotators were native English speakers, who each completed the same randomly chosen 300 example subset of the development set. The examples were sampled such that half (150) of these examples came from the few-shot split and the other half came from the zero-shot split. Similar to the modeling baselines, we evaluate each annotator's performance by scoring them against the sense associated with that example in Wiktionary (which we assume to be gold labels).

\section{Baseline Experiments}
\subsection{Experimental Setup}
\label{baseline-exp-setup}
\paragraph{Data} All baselines for the FEWS dataset are trained on the \textit{train} set unless specifically stated to have been trained on the \textit{extended train} set. All models for FEWS are tuned using the \textit{development} set and then evaluated on the held-out \textit{test} set.

\paragraph{Experimental Details}
Our probe and BEM baselines are in implemented in PyTorch\footnote{https://pytorch.org/} and optimized with Adam \cite{kingma2015adam}. For the BEM, we use the implementation provided by \citet{blevins2020moving}.\footnote{https://github.com/facebookresearch/wsd-biencoders} We obtain the \textit{bert-base-uncased} encoder from \citet{Wolf2019HuggingFacesTS} to get the BERT output representations for the probe and to initialize the BEM models. Further hyperparameter details are given in Appendix~\ref{appendix-parameters}. 

\subsection{Modeling Results}
Table \ref{fews-results-table} shows the results of our baseline experiments on FEWS. We find that the MFS baseline is weak overall, primarily because it is unable to predict anything about the held-out, zero-shot senses; the Lesk algorithms both outperform this baseline in the overall setting, with the Lesk+emb approach scoring slightly better than the original Lesk approach by 1.78-4.2\% across the different evaluation subsets. However, on the few-shot examples in both the development and test sets, we see that the MFS baseline outperforms both of the Lesk baselines. This shows that, for the few-shot examples, the MFS heuristic remains a reasonably strong baseline even with the more uniform sense distribution of FEWS (and indicates that the distribution of examples drawn from the dictionary is less uniform than expected).

The neural baselines we run generally outperform the knowledge-based ones. The \textbf{Probe$_{BERT}$} model does fairly well on the few-shot examples, outperforming the MFS baseline by about 20 accuracy points; however, it is unable to disambiguate words in the zero-shot splits correctly since the probe can not generalize to unseen senses. In comparison, \textbf{BEM$_{BERT}$} performs well across the entire evaluation set. In particular, the BEM achieves much better zero-shot performance than other baselines, though performance on this subset still lags behind few-shot performance. Finally, we see that humans perform better than all of the considered baselines, particularly on zero-shot senses where humans outperform the BEM by 11.53 points. More details about the human evaluation are given below (Section \ref{human-eval-sec}).

Additionally, we find that training on the extended train set has little effect on these baselines: the \textbf{MFS} and \textbf{Probe$_{BERT}$} baselines perform slightly worse (with deltas of \mbox{-0.08\%} and -0.21\% on the test set, respectively), and \textbf{BEM$_{BERT}$} performs 1.05\% better. Appendix~\ref{appendix-extended_train} presents full results for the extended train baselines.

\begin{table}[t]
    \centering
    \begin{tabular}{l | r r r } 
    \toprule
     & \multicolumn{3}{c}{\textbf{Dev Subset}} \\
     & \textbf{Full Set}& \textbf{Few-shot} & \textbf{Zero-shot} \\
    \hline
    \hline
    \multicolumn{4}{l}{\textbf{Knowledge-based baselines}} \\
    \hline
    MFS & 23.66 & 47.33 & 0.00\\
    Lesk & 38.33 & 41.33 & 35.33 \\
    Lesk+emb & 46.00 & 48.67 & 43.33 \\
    \hline
    Human & 80.11 & 80.44 & 79.87 \\
    \hline
    \hline
    \multicolumn{4}{l}{\textbf{Neural baselines}} \\ 
    \hline
    Probe$_{BERT}$ & 33.67 & 67.33 & 0.00 \\
    BEM$_{BERT}$ & 73.00 & 80.66 &65.33 \\
    \toprule
    \end{tabular}
    \caption{Accuracy (\%) of our baselines on the subset of the development set manually scored by human annotators.}
    \label{human-subset-baselines}
\end{table}

\subsection{Human Evaluation Results}
\label{human-eval-sec}
The human annotators achieved an average accuracy of 80.11\% (with each annotator getting 84.67\%, 78.67\%, and 77.00\% accuracy) and an average inter-annotator agreement of $\kappa = 0.802$. We find that humans perform slightly better on the examples that correspond to few-shot examples for the dataset than those corresponding to zero-shot examples despite not using the training data, with an average of 80.44\% and 79.87\% accuracy on those two subsets, respectively.

We also report the performance of the baselines for FEWS on the same subset of the development set that was manually completed by humans (Table \ref{human-subset-baselines}). We find that the baselines perform similarly on this subset, with a small decrease in performance compared to the full development set (with decreases in accuracy ranging between 0.81\% and 3.46\% when compared to the development set).

\begin{table*}[t]
    \centering
    \begin{tabular}{l | c || c c c c || c c c c | c}
    \toprule
     & \textbf{Dev} & \multicolumn{4}{c ||}{\textbf{Test Datasets}} & \multicolumn{5}{c} {\textbf{Concatenation of all Datasets}} \\
    & SE07 & SE2 & SE3 & SE13 & SE15 & Nouns & Verbs & Adj. & Adv. & ALL \\
    \hline
    MFS (in train data) & 54.5 & 65.6 & 66.0 & 63.8 & 67.1 & 67.7 & 49.8 & 73.1 & 80.5 & 65.5 \\
    BEM$_{BERT}$ & \textbf{74.5} & \textbf{79.4} & 77.4 & \textbf{79.7} & \textbf{81.7} & \textbf{81.4} & 68.5 & \textbf{83.0} & 87.9 & \textbf{79.0} \\
    BEM$_{FEWS}$ & 73.6 & 79.1 & \textbf{77.9} & 79.1 & 81.6 & 81.2 & \textbf{68.9} & 81.8 & \textbf{88.2} & 78.8\\
    \hline
    BEM$_{zero-shot}$ & 53.0 & 66.6 & 62.3 & 69.1 & 74.9 & 70.7 & 51.2 & 72.0 & 69.7 & 66.4\\
    \toprule
    \end{tabular}
    \caption{F1-score on the English all-words WSD in the WSD Evaluation Framework \cite{raganato2017word}. We compare the best model from \citet{blevins2020moving} (BEM$_{BERT}$) against (1) a model first trained on FEWS and then trained on SemCor (BEM$_{FEWS}$) and (2) a model trained on FEWS and evaluated on this task without further finetuning (BEM$_{zero-shot}$).}
    \label{semcor-results-table}
\end{table*}

\begin{table}[t]
\centering
    \resizebox{\columnwidth}{!}{
    \begin{tabular}{l| c c | c c }
    \toprule
    & \multirow{2}{*}{\textbf{MFS}} & \multirow{2}{*}{\textbf{LFS}} & \multicolumn{2}{c}{\textbf{Zero-shot}}\\
    & & & \textbf{Words} & \textbf{Senses} \\
    \hline
    WordNet S1 & 100.0 & 0.0 & 84.9 & 53.9 \\
    BEM$_{BERT}$ & \textbf{94.1} & 52.6 & 91.2 & 68.9 \\
    BEM$_{FEWS}$ & 93.7 & 52.9 & 92.2 & 74.8\\
    \hline
    BEM$_{zero-shot}$ & 72.6 & \textbf{55.5} & \textbf{92.7} & \textbf{80.5} \\ 
    \toprule
    \end{tabular}}
    \caption{F1-score on the MFS, LFS, and zero-shot subsets of the \textbf{ALL} evaluation set from the WSD Evaluation Framework. Zero-shot examples are the words and senses (respectively) from the evaluation suite that do not occur in SemCor.}
    \label{semcor-lfs-table}
\end{table}

\section{Transfer Learning with FEWS}
\label{transfer-section}

Next, we investigate how useful FEWS is at improving WSD performance on existing benchmarks. We perform transfer learning experiments by iteratively finetuning models on FEWS and the WSD Evaluation Framework \cite{raganato2017word}, with one acting as the \textit{intermediate} dataset and evaluating performance on the other, \textit{target} dataset. We find that on global metrics, this approach performs similarly to finetuning only on the training data for each benchmark; however, transferring from FEWS to the WSD Framework improves performance on less-frequent and zero-shot senses. This suggests that FEWS provides valuable WSD information not covered in SemCor.

\subsection{Experimental Setup}

We apply the supplementary training approach presented in \citet{phang2018sentence} to perform our transfer learning experiments.  We initialize a BEM with the best model developed on the \textit{intermediate} dataset and evaluate on the \textit{target} dataset in two ways: first, by evaluating this BEM on the \textit{target} dataset in a zero-shot manner, without additional finetuning; and second by finetuning the BEM on the \textit{target} training data before performing the \textit{target} evaluation. We refer to these models as \textbf{BEM$_{zero-shot}$} and  \textbf{BEM$_{intermediate}$}, respectively. As a baseline, we also compare against the best BEM trained only on the \textit{target} dataset (with no exposure to the \textit{intermediate} dataset), \textbf{BEM$_{BERT}$}.\footnote{We note that this is a naive approach to transferring between Wiktionary senses and WordNet synsets, and that it is possible that better transfer learning could occur with a more complicated multitask approach or by mapping between the two lexical resources with approaches such as \citet{navigli2012babelnet} or \citet{miller2014wordnet}.} The model implementation details are identical to those for the baseline experiments on FEWS (Section \ref{baseline-exp-setup}).

\subsection{Data}
Models that are finetuned for the WSD Evaluation Framework are trained using SemCor, a large dataset annotated with WordNet synsets and commonly used for training WSD models \cite{miller1993semantic}. Following previous work, we use SemEval-2007 as a validation set (\textbf{SE07}; \cite{pradhan2007semeval}) and hold out the other evaluations sets in the Framework (Senseval2 (\textbf{SE2}; \cite{palmer2001english}), Senseval-3 (\textbf{SE3}; \cite{snyder2004english}), SemEval-2013 (\textbf{SE13}; \cite{navigli2013semeval}), and SemEval-2015 (\textbf{SE15}; \cite{moro2015semeval}) as test sets. Similarly to the baseline experiments, models that are trained on FEWS use the \textit{train} set (note that we do not use the \textit{extended train} set in these experiments), are validated on the \textit{development} set, and finally evaluated on the held out \textit{test} set.

\subsection{Results}
\paragraph{FEWS Results} We first consider the setting where SemCor acts as the \textit{intermediate} dataset and FEWS as the \textit{target} (Table \ref{fews-results-table}).
We find that \textbf{BEM$_{SemCor}$} performs similarly to training only on FEWS (with an improvement of 0.21\% on the test set). \textbf{BEM$_{zero-shot}$}, which is not finetuned on the FEWS train set, unsurprisingly performs worse than any of the BEMs that saw the FEWS training data but outperforms the Lesk baselines.

\paragraph{WSD Evaluation Framework Results} We then consider the opposite setting, where FEWS is the \textit{intermediate} dataset and the WSD Evaluation Framework acts as the \textit{target} evaluation (Table \ref{semcor-results-table}). On the overall evaluation set, we again see that \textbf{BEM$_{FEWS}$} performs similarly to the \textbf{BEM$_{BERT}$} baseline on the overall \textbf{ALL} metric, and that the zero-shot BEM model underperforms the other biencoders. 

We then break down performance on the \textit{target} evaluation set by sense frequencies: we evaluate performance on the most frequent sense (MFS) of each word in the evaluation (i.e., the sense each word takes most frequently in the SemCor training set); the less frequent senses (LFS) of words, or any sense a words takes besides its MFS; zero-shot words that are not seen during training on SemCor; and zero-shot senses, also not seen during training (Table \ref{semcor-lfs-table}).  
We find that both transfer learning models perform better on LFS and zero-shot examples than \textbf{BEM$_{BERT}$}.\footnote{For the models trained on FEWS, it is possible they have seen closely related senses to those in the zero-shot subsets of from the Wiktionary sense inventory; however, these are represented differently in each dataset and correspond to different definitions to be encoded by the biencoder.}

In particular, the zero-shot transfer model does well on these subsets, demonstrating that a fair amount of WSD knowledge about uncommon senses can be transferred in a zero-shot manner between these datasets (albeit at the expense of higher performance on the MFS subset). This result also shows how much the MFS group dominates existing WSD metrics and highlights the need for focused evaluations of other types of word senses. Finally, we see that even without exposure to the natural sense distribution in natural language texts, the zero-shot model still performs significantly better on the MFS of words than the LFS, with a 17.1 F1 point difference between the two subsets; this is likely because the BERT encoder is exposed to the sense distribution of English natural language documents during pretraining.

\section{Conclusion}
We establish baseline performance on FEWS with both knowledge-based approaches and recently published neural WSD models. Unsurprisingly, neural models based on pretrained encoders perform best on FEWS; however, the human evaluation shows there is still room for improvement, particularly for zero-shot senses. Finally, we also present results on transferring word sense knowledge from FEWS onto existing WSD datasets with staged finetuning. While our naive approach for transferring knowledge from FEWS does not improve performance on the global WSD metric, adding FEWS as an additional training signal improves performance on uncommon senses in existing evaluation sets. 

We hope that FEWS will inspire future work focusing on better methods for capturing rare senses in WSD and better modeling of word sense in niche domains like internet culture or technical writing.  

\section*{Acknowledgements}
This material is based on work conducted at the University of Washington, which was supported by the National Science Foundation Graduate Research Fellowship Program under Grant No. DGE-1762114. The authors thank Julian Michael, Maarten Sap, and the UW NLP group for helpful conversations and comments on the work.

\bibliography{anthology,eacl2021}
\bibliographystyle{acl_natbib}

\begin{table*}[t]
    \centering
    \begin{tabular}{l r| r r r | r r r} 
    \toprule
     & & \multicolumn{3}{c|}{\textbf{Dev}} & \multicolumn{3}{c}{\textbf{Test}} \\
     & & \textbf{Full Set}& \textbf{Few-shot} & \textbf{Zero-shot} & \textbf{Full Set} & \textbf{Few-shot} & \textbf{Zero-shot} \\
    \hline
    \hline
    \multirow{3}{*}{MFS} & Ext. Train & 26.13 & 52.26 & 0.00 & 25.65 & 51.30 & 0.00 \\
    & Train & 26.39 & 52.78 & 0.00 & 25.73 & 51.46 & 0.00 \\
    & $\Delta$ & -0.26 & -0.52 & -- & -0.08 & -0.16 & -- \\
    \hline
    \multirow{3}{*}{Probe$_{BERT}$} & Ext. Train & 36.03 & 72.06 & 0.00 & 35.86 & 71.72 & 0.00 \\
    & Train & 36.17 & 72.34 & 0.00 & 36.07 & 72.14 & 0.00 \\
    & $\Delta$ & -0.14 & -0.28 & -- & -0.21 & -0.42 & -- \\
    \hline
    \multirow{3}{*}{BEM$_{BERT}$} & Ext. Train & 74.12 & 79.38 & 68.86 & 73.82 & 79.70 & 67.94\\
    & Train & 73.81 & 79.28 & 68.34 & 72.77 & 79.06 & 66.48 \\
    & $\Delta$ & 0.31 & 0.10 & 0.52 & 1.05 & 0.64 & 1.46 \\
    \toprule
    \end{tabular}
    \caption{Accuracy of the FEWS baselines trained on the extended train set. In each group of rows, we report (1) the extended train baseline, (2) the comparable baseline trained on the standard train set, and (3) the performance delta between the two (where a positive delta indicates the extended train baseline performs better).}
    \label{extended-train-baselines}
\end{table*}

\newpage
\appendix

\section{Model Hyperparameters}
\label{appendix-parameters}
Each model reported in this paper was tuned on a single hyperparameter sweep over the reported ranges and chosen based on the appropriate development set metric (accuracy on FEWS, F1 performance on the Unified WSD Framework). 

\paragraph{Probe$_{BERT}$} The linear layer in the BERT probe baseline is trained for 100 epochs. It is tuned over a range learning rates ($[5e-6, 1e-5, 5e-5, 1e-4]$, with a final learning rate of $1e-4$). We use a batch size of 128 to train this probe. 

\paragraph{BEM} For the biencoder model (BEM), we use the codebase provided by \cite{blevins2020moving}. Following this work, we train the BEM for 20 epochs with a warmup phase of 10,000 steps; we use a context batch size of 4 and a gloss batch size of 256. Each BEM is tuned over the following learning rates: $[1e-6, 5e-6, 1e-5, 5e-5]$. The BEM$_{BERT}$ and BEM$_{SemCor}$ had a final learning rate of $5e-6$, and the BEM$_{FEWS}$, of $1e-6$.

\section{Extended Train Baselines}
\label{appendix-extended_train}
The extended train set in FEWS contains all of the quotation-based examples from the train set as well as the additional, shorter example \textit{illustrations} that are written by Wiktionary editors to exemplify a particular sense of a word. We retrain the \textbf{MFS}, \textbf{Probe$_{BERT}$}, and \textbf{BEM$_{BERT}$} baselines on this extended training set; the other baselines we consider (\textbf{Lesk} and \textbf{Lesk+emb}) are calculated without using either of the training sets.

Table \ref{extended-train-baselines} compare the extended train baselines against those trained on the standard train set. For the \textbf{MFS} and \textbf{Probe$_{BERT}$} baselines, we find that adding the extra, stylistically different  \textit{illustrations} in extended train slightly hurts performance. However, the stronger \textbf{BEM$_{BERT}$} is able to better use this data and achieves somewhat stronger performance with additional training data. Notably, most of this improvement in the \textbf{BEM$_{BERT}$} comes from the zero-shot evaluation setting, even though the extended train set does not contain any of these zero-shot senses.

\end{document}